\pdfoutput=1
\documentclass[11pt]{article}

\usepackage[final]{acl}
\usepackage{multirow}% http://ctan.org/pkg/multirow
\usepackage{hhline}
\usepackage{makecell}
\usepackage{amssymb}% http://ctan.org/pkg/amssymb
\usepackage{pifont}% http://ctan.org/pkg/pifont
\newcommand{\cmark}{\ding{51}}%
\newcommand{\xmark}{\ding{55}}%
\usepackage[colorinlistoftodos,prependcaption,textsize=tiny]{todonotes}
\usepackage{caption}
\usepackage{soul}
\usepackage{times}
\usepackage{latexsym}

\usepackage{algorithm}
\usepackage{algorithmic}
\usepackage{amsmath}

\usepackage{tikz}
\usetikzlibrary{shapes,arrows}

\usepackage[T1]{fontenc}
\usepackage[utf8]{inputenc}

\usepackage{microtype}

\usepackage{inconsolata}

\usepackage{graphicx}

\newcommand*\samethanks[1][\value{footnote}]{\footnotemark[#1]}

\def\W{\mathbf{W}}

\def\real{\rm I\!R}

\def\U{\mathbf{U}}
\def\u{\mathbf{u}}
\def\V{\mathbf{V}}
\def\v{\mathbf{v}}
\def\SSigma{\boldsymbol{\Sigma}}
\usepackage{amsmath}
\title{Single Parent Family: A Spectrum of Family Members from a Single Pre-Trained Foundation Model}

\author{
Habib Hajimolahoseini\thanks{Contributed equally; authorship order determined randomly.},
Mohammad Hassanpour\samethanks[1]\thanks{Work done while at Huawei Technologies.},
Foozhan Ataiefard,\\
\textbf{Boxing Chen,
Yang Liu}\\
Toronto Research Centre, Huawei Technologies\\
\href{mailto:hassanpour71@gmail.com}{\texttt{\textcolor{black}{hassanpour71@gmail.com}}}
}
\begin{document}
\maketitle
\begin{abstract}
This paper introduces a novel method of Progressive Low Rank Decomposition (PLRD) tailored for the compression of large language models. 
Our approach leverages a pre-trained model, which is then incrementally decompressed to smaller sizes using progressively lower ranks. 
This method allows for significant reductions in computational overhead and energy consumption, as subsequent models are derived from the original without the need for retraining from scratch. 
We detail the implementation of PLRD, which strategically decreases the tensor ranks, thus optimizing the trade-off between model performance and resource usage. 
The efficacy of PLRD is demonstrated through extensive experiments showing that models trained with PLRD method on only $\textbf{1B}$ tokens maintain comparable performance with traditionally trained models while using $\textbf{0.1\%}$ of the tokens. The versatility of PLRD is highlighted by its ability to generate multiple model sizes from a single foundational model, adapting fluidly to varying computational and memory budgets. 
Our findings suggest that PLRD could set a new standard for the efficient scaling of LLMs, making advanced AI more feasible on diverse platforms.
\end{abstract}

\section{Introduction} \label{introduction}
\begin{figure}[t]
  \includegraphics[scale=0.52]{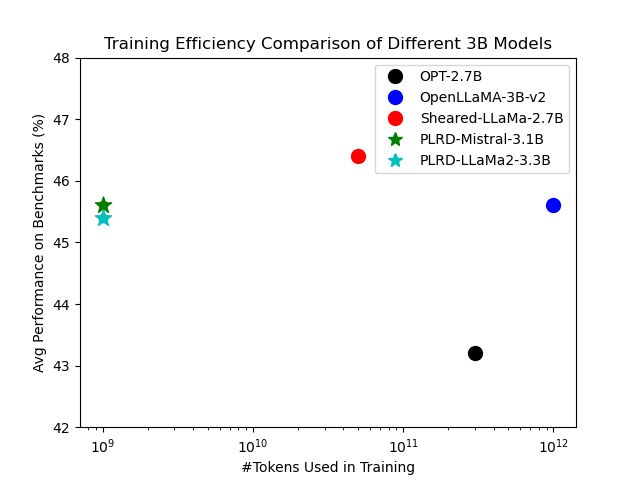} 
  \caption {Training Efficiency Comparison of 3B Models. PLRD-Mistral-3.1B and PLRD-LLaMa2-3.3B models achieve similar average performance on downstream benchmarks while they are trained with 0.1\% of number of tokens used in pre-training from scratch.}
\end{figure}
\begin{table*}
  \centering
  \begin{tabular}{cccc}
    \hline
    \textbf{Model} & \textbf{\# of Training Tokens} & \textbf{Loss-based Data Selection} & \textbf{Extra Training Stage} \\
    \hline
    OPT-2.7B & 300B & \xmark & \xmark \\
    Open-LLaMA-3B-v2 & 1T& \xmark & \xmark \\
    Sheared-LLaMA-2.7B & 50B & \cmark & \cmark \\
    PLRD-Mistral-v0.1-3.1B & 1B & \xmark & \xmark \\

    \hline
  \end{tabular}
  \caption{\label{citation-guide}
    Comparison of PLRD, LLaMa-Shearing \cite{xia2023sheared}, and pretrain from Scratch methodologies. PLRD is effective without the need for large number of tokens in training or specific training recipe to achieve similar results.
  }
\end{table*}
Recent advancements in deep learning have led to the development of increasingly large models where the parameter count often exceeds several billion \citep{zhao2023survey}. 
Large Language Models (LLMs) typically use float32 or float16 formats, where each float16 takes up 2 bytes. Thus, a model, for example, GPT-3 with 175B parameters requires 320 gigabytes, making it too large for most consumer devices (it requires at least five A100 GPUs only for running inference) \citep{zhu2023survey}.

LLMs are often released as a series of variants or family members with different sizes to accommodate a broad spectrum of computational resources and application needs. 
For instance, the Llama2 model is available in different sizes including 7 billion, 13 billion and 70 billion parameters. 
This tiered approach allows developers and organizations to select a model size that best fits their specific performance requirements and budget constraints. 
However, the problem is that all of these family members are trained from scratch and therefore the number of family members is very low (in order of 3-4). 
Even the smaller LLMs e.g. TinyLlama \citep{zhang2024tinyllama}, and Phi-2 \citep{javaheripi2023phi}, are trained from scratch with billions of tokens which requires significant computation and memory. 
Furthermore, if the computational budget of a user allows a model size in between the two family members, the user has to pick the family member with the smaller size that may not be optimum for their use case. 
% Pre-training, fine-tuning or inference on these large scale models requires vast amounts of computational resources and memory, making them impractical for deployment on resource-constrained environments such as mobile devices or edge computing platforms. 

Model compression techniques, address these issues by reducing the model's size and complexity without significantly compromising its performance. 
This reduction not only enhances the feasibility of deploying advanced AI capabilities in everyday applications but also democratizes access to state-of-the-art technologies, making them accessible to a broader range of users and developers \citep{dean2012large}. 
Model compression techniques can be classified into four major categories including: Quantization, Pruning, Knowledge Distillation and Low-Rank Factorization \citep{cheng2017survey, zhu2023survey}. 

In contrast with using floating point numbers to represent the model weights or activations, quantization techniques simplify these into integers or other lower precision units, thereby lowering both storage needs and computational complexity \citep{bie2019simplified}. 
If quantization is applied during training or fine-tuning, this is called Quantization-Aware Training (QAT) \citep{ding20224}.
It may be applied after training is finished in which case it is known as Post-Training Quantization (PTQ) \citep{liu2021post}. 

Pruning is another technique used to compress deep learning models that eliminates unnecessary or redundant parameters that do not significantly impact the performance of the model \citep{lebedev2016fast, wen2016learning, fang2023depgraph}. 
% This process enhances the model's storage and memory efficiency, and reduces computational demands, making the model more practical for deployment in resource-limited environments. 
Pruning is categorized into two types: structured and unstructured. Structured pruning focuses on removing entire units or layers based on predefined criteria to maintain a coherent network structure, while unstructured pruning targets individual weights, leading to a sparsely connected network. 

Knowledge Distillation (KD) improves model performance by transferring knowledge from a complex "teacher" model to a simpler "student" model \citep{hinton2015distilling}. 
It uses two approaches: Black-box KD, where only the teacher’s predictions are used, and White-box KD, which also accesses the teacher's internal parameters \citep{mirzadeh2020improved}. 

Low-Rank Factorization is a model compression strategy used to reduce the dimensions of weight matrices in neural networks by decomposing a large matrix into two or more smaller matrices. 
This factorization leads to a significant decrease in the number of parameters and computational requirements. 
Tensor Train Decomposition (TTD) and Singular Value Decomposition (SVD) are two well known techniques used for decomposing weight matrices.
Note that although techniques such as Low Rank Adaptation (LoRA) and its variants are highly adapted to fine-tune LLMs efficiently, they are not considered model compression techniques, as the number of parameters of the fine-tuned and original is exactly the same \citep{hu2021lora, hajimolahoseini2023training}. 

Despite the widespread use and some advantages of model compression techniques, they have some limitations in practice: pruning can lead to irregular sparsity and requires extensive fine-tuning, quantization risks significant precision loss and relies on specific hardware support, and knowledge distillation involves additional training overhead \cite{xia2023sheared, xia2022structured}. 
The main advantage of Low-Rank Decomposition techniques is that they do not necessitate extensive pre-training since they use a large pre-trained model to initialize the compressed model. 
Consequently, LRD often needs only minimal fine-tuning to regain most of the accuracy lost during compression.
Furthermore, in contrast with most of the model compression techniques, the user has control over the compression ratio and so, one can compress large language models according to their exact computation and memory budget. 

However, it's important to recognize that for substantial compression factors, LRD may result in a poor approximation of the original weights, making it challenging to restore accuracy through fine-tuning alone. 
\begin{table*}
  \centering
  \begin{tabular}{cccccccccc}
    \hline
    \textbf{Model} \small{(\#Tokens to Train)} & \textbf{LogiQA} & \textbf{BoolQ} & \textbf{MMLU}&  \textbf{WinoGrande} & \textbf{Average}  \\
    % \hline
    % Mistral-v0.1-7B & 30.3\% & 83.6\% & 59.6\% & 74.0\% & 61.9\% \\  
    % LLaMA2-7B & 30.3\% & 77.8\% & 41.9\% & 69.0\% & 54.7\% \\
    \hline
    OPT-2.7B \small{(300B)} & 25.8\% & 60.3\% & 25.6\% &60.9\% & 43.1\% \\
    Open-LLaMA-3B-v2 \small{(1T)} & 28.0\% & 65.5\% & 25.5\% & 63.3\% & 45.5\% \\
    Sheared-LLaMA-2.7B \small{(50B)} & $\textbf{28.9\%}^\ast$ & \textbf{66.0\%} & \textbf{26.5\%} & $\textbf{64.2\%}^\ast$ & \textbf{46.4\%} \\
    \hline
    % PLRD-Mistral-v0.1-5.2B & 26.4\% & 72.6\% & 30.4\% & \66.8\% & 49.0\%\\
    % PLRD-Mistral-v0.1-4.9B & 28.3\% & 69.6\% & 30.9\% & 63.3\% & 48.0 \\
    % PLRD-Mistral-v0.1-4.1B & 28.9\% & 68.6\% & 27.3\% & 62.1\% & 46.7\%\\
    PLRD-Mistral-v0.1-3.1B \textbf{(Ours)} \small{(1B)} & 26.9\% & \textbf{66.0\%} & \textbf{26.5\%} & 63.1\% & \underline{45.6\%}\\
    PLRD-LLaMa2-3.3B \textbf{(Ours)} \small{(1B)} & \underline{28.3\%} & 64.1\% & 25.7\% & 63.4\% & 45.4\%\\
    \hline
    
  \end{tabular}
  \caption{\label{benchmark-3b}
    Accuracy of models on a variety of different benchmarks are evaluated in zero-shot setting. The best result in each benchmark is \textbf{bold} and the second best is \underline{underlined}.\\
    $^\ast$ Accuracy of Sheared-LLaMa-2.7B on LogiQA and WinoGrande are directly adopted from Sheared-LLaMa paper \cite{xia2023sheared}. 
  }
\end{table*}
To mitigate this issue, we propose progressive LRD, which applies compression iteratively in small steps, with each step building on top of the previous model, thus improving the initialization point on the loss surface.
This allows us to start from a large pre-trained model and compress it continuously down to any size that satisfies our computation and memory budget.
Furthermore, it provides a spectrum of models with an unlimited number of family members for each pre-trained LLM without needing to pretrain from scratch. 

% While LRD has been employed to compress large transformer-based models in previous research, these studies typically use a single-shot approach, where all layers are decomposed simultaneously before fine-tuning. 
% This could result in a drastic drop in model performance. 
% There is a lack of comprehensive studies comparing different LRD strategies, such as layer-by-layer and progressive decomposition, specifically for transformer-based models. 
% This paper explores the application of LRD using various strategies to assess their effectiveness in the extreme compression of transformer-based models.

\begin{figure*}[t]
  \includegraphics[width=0.48\linewidth]{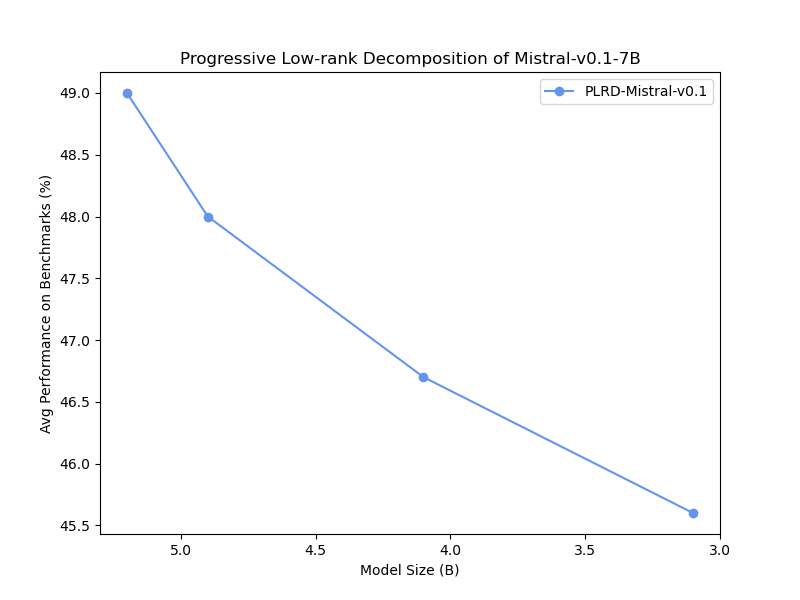} \hfill
  \includegraphics[width=0.48\linewidth]{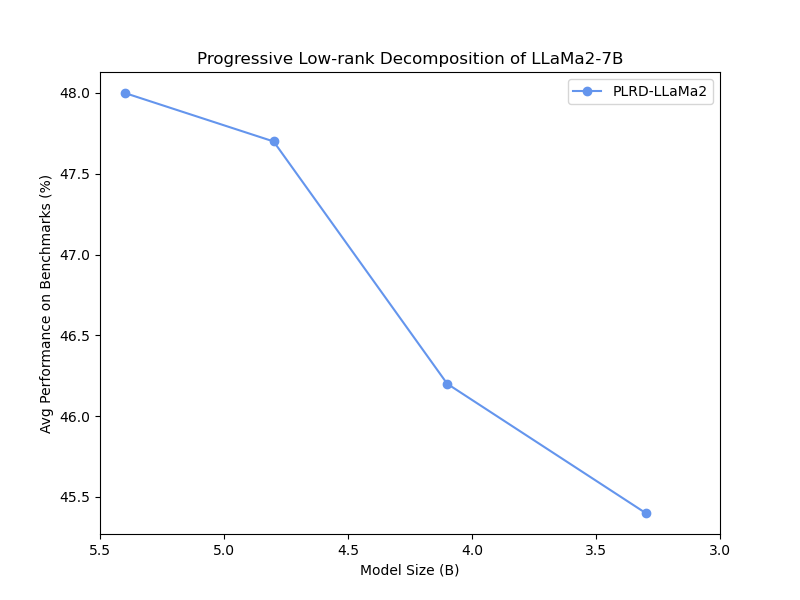}
  \caption {Progressive Low-rank Decomposition of Mistral-v0.1 (left) and LLaMa2 (right). This figure shows the steps of compression and training for each of these two models. In each step of compression, the accuracy of the model was recovered with continual pretraining on 250M tokens. The final models are trained on 1B token in total.}
\end{figure*} 
% \section{Related Work}
% \section{Methodology}
% In this section, the two main parts of PLRD are explained. First, we delve into Low Rank Factorization as a mathematical backbone of PLRD. Then, the second part lays down the foundations of progressive decomposition approach that uses low rank factorization to compress LLMs.
\section{Low Rank Factorization} \label{low-rank}

Large language models are a stack of transformer layers, in which the fully connected (FC) layers are the building blocks of the feed forward and self-attention modules \citep{vaswani2017attention}. 
For simplicity in representations, we assume these FC layers have weight matrices of shape $\W\in \real^{d_{in}\times d_{out}}$, where $d_{in}$ and $d_{out}$ represent the number of input and output features, respectively. 
In this case, the number of parameters in this FC layer could be calculated as: ${d_{in}\times d_{out}}$.

The rank of matrix $\W$ can be interpreted as the number of its linearly independent columns or rows.
% sometimes even the condition index, condition number, normalized trace or determinants are used as proxy to the rank.  
Singular Value Decomposition (SVD), is a common tool used for factorizing a matrix into its components. 
% This optimization problem has an analytical solution as follows \citep{van1987matrix}:
Assume that the weight matrix $\W\in \real^{d_{in}\times d_{out}}$ is decomposed using SVD as follows
\begin{equation}\label{svd}
W = \U\SSigma \V^\top=\sum_{i=1}^r{\sigma_i\u_i\v_i^\top},
\end{equation}
where $\U \in \real^{d_{in}\times d_{in}}$ and $\V \in \real^{d_{out}\times d_{out}}$ are the orthogonal matrices and $\SSigma \in \real^{d_{in}\times d_{out}}$ is a diagonal rectangular matrix containing the singular values $\sigma_i>0$ of $\W$ with rank $r$ ($r$ is the number of non-zero singular values).
In \eqref{svd}, if we only use the first $R$ terms of the summation, the resulted matrix $\W'$ would be an approximation of $\W$ with a lower rank $R<r$:
\begin{equation}\label{svd2}
\W' = \sum_{i=1}^R{\sigma_i\u_i\v_i^\top}=\U'\SSigma' \V'^\top
\end{equation}
where $\U' \in \real^{d_{in}\times R}$ and $\V' \in \real^{d_{out}\times R}$ are the new  orthogonal matrices and $\SSigma' \in \real^{R\times R}$ is the new diagonal rectangular matrix.
% Therefore, the low rank estimation  problem could be explained as the following minimazation
% \begin{equation}\label{svd3}
% \min_{\mathrm{rank}(\X) \leq R} \norm{\W-\X} = \norm{\W-\W'}.
% \end{equation}
Based on \eqref{svd2}, $\W'$ could be represented as an inner product of two matrices $\W_0$ and $\W_1$ i.e.
\begin{align}
\W' = \W_0 \W_1, \label{lrd-mul}
\end{align}
where:
\begin{align}
\quad \W_0 &= \U'\sqrt{\SSigma'}, \quad \\
\W_1 &= \sqrt{\SSigma'} \V'^\top \label{svd4}
\end{align}
where $\W_0 \in \real^{d_{in}\times R}$ and $\W_1 \in \real^{R\times d_{out}}$, and $\sqrt{\SSigma}$ is a diagonal matrix of square root of singular values $\sqrt{\sigma_i}.$ 
If we replace the FC layer $\W$ with the two consecutive FC layers $\W_0$ and $\W_1$, the number of parameters could shrink significantly depending on the rank $R$, where $R$ controls the compression ratio.
The compression ratio CR can be calculated as follows:
\begin{align}
\text{CR} = \frac{d_{in} \times d_{out}}{R \times (d_{in}+d_{out})}
\label{ratio_selection}
\end{align}
where $1\leq R\leq \min(d_{in}, d_{out})$ \cite{hajimolahoseini2023training}.

If $R$ is too large, the product of decomposed matrices $\W_0$ and $\W_1$ is closer to the original matrix $\W$ and the drop in accuracy is negligible with the price of small compression or even no compression.  
% The square $\SSigma $of size $R\times R$ fits within the rectangle $\W$ of size $C\times  S$,  if $R\leq \min(C,S)$, which is a standard result in linear algebra.

%\subsubsection{Rank Estimation}
% Different approaches  are proposed to  estimating the rank $R$ in practice. A simple approach is to specify the rank  based on the desired compression ratio for each FC layer. Thus, using \ref{num_param2} if we want the FC layer to be compressed by a factor of $P$, the rank $R$ is $R = \frac{C S}{P (C+S)}.$
% The range of the rank  $R$ is quite important. If $R$ is large, the product of decomposed matrices $\W_0$ and $\W_1$ would be closer to the original matrix $\W$ and the drop in accuracy is negligible with the price of small compression or even no compression.  The square $\SSigma $of size $R\times R$ fits within the rectangle $\W$ of size $C\times  S$,  if $R\leq \min(C,S)$, which is a standard result in linear algebra.

On the other hand, if $R$ is too small, the approximation in \eqref{svd2} is inaccurate and accuracy loss is  significant with the advantage of high compression ratio. 
In the extreme case of $R=1$, the matrix $\SSigma'$ becomes an scalar and thus, $\U'$ and $\V'$ reduce to vectors.
% Therefore, the upper and lower bounds of the rank $R$ and their respective compression ratio $P$ is 
% \begin{align}
%  \frac{C S}{\min(C,S)(C+S)} \leq P \leq \frac{C S}{C+S}.
% \end{align}

% Define a hidden node of a single FC layer as 
% \begin{eqnarray*}
%         f_\W(x) = \w_2 ^ \top a\left(\W\x\right),
% \end{eqnarray*}
% which is used to approximate a continuous function $f(\x)$ on a compacta $K\subset \real^C,$ in which $a(\cdot)$ is an activation function.
% The following theorem shows under some bound on singular values, a low rank FC layer  behaves similar to a full rank layer. 
% \begin{theo}
%  A FC layer of rank ${R}$ and $n$ hidden units defined on a compacta $K\subset \real^C$ with $L$-Lipschitz activation is can be approximated any continuous  $f\in C(K)$ with error $(\epsilon+\delta)$, for a large enough $n$ given 
%  $$\sigma_{r+1} < \delta L^{-{1\over 2}}( \norm K  \norm{\w_2})^{-1}$$
%  where $\sigma_{r+1}$ is the $r+1$ singular value of the original weight matrix  $\W$,   $\epsilon$ is the approximation quality of the full rank neural network $\norm{f(\x) - f_\W(\x)}<\epsilon,$ and $\delta$ is the approximation error added due to the low rank decomposition.
% \end{theo}
% See Appendix for the proof.

\section{Progressive Decomposition} \label{pd}
In contrast with most of the literature that applies low rank factorization to all layer in a single-shot manner, we decompose the models in multiple progressive steps using a progressively decreasing rank in each step.
Each compression step is followed by a fine-tuning stage to make sure the acuuracy is recovered before going to the next step. 
% It could be applied to all layers of the model at the same time, or it could be done layer by layer. 
% However, due to the aggressive target compression ratio, we decompose the NLP models in multiple intermediate steps (applying SVD to all layers in case of LRD) instead of one-shot compression, where we follow each compression step with some finetuning epochs.
% This method is actually inspired by the method proposed in \citep{gusak2019automated} for the convolutional networks.
% To this end, we use a progressive rank selection approach in which, we start from larger values for rank and decompose the entire model altogether with a low compression ratio.
% After applying each LRD stage, the accuracy will drop.
% Therefore, the model is fine-tuned for few epochs in order to recover the accuracy. 
% The rank $R$ is decreased in each stage so that we have a higher compression ratio. 

In order to avoid doubling the number of layers after each compression step, we only decompose the second decomposed matrix $\W_1$ into $\W_1^0$ and $\W_1^1$ and multiply the matrices $\W_0$ and $\W_1^0$. So according to \eqref{lrd-mul}, we have:
\begin{align}
\W &\xrightarrow{\text{LRD}} \W_0, \W_1 \\ 
&\xrightarrow{\text{Apply LRD to $\W_1$}} \W_0, (\W_1^0, \W_1^1)\\
&\xrightarrow{\text{Multiply $\W_0$, $\W_1^0$}} (\W_0\W_1^0), \W_1^1
% \W_1 &\xrightarrow{\text{LRD}} \W_1^0, \W_1^1 \\
% \W &\xrightarrow{} (\W_0 \times \W_1^0), \W_1^1
\label{prog-svd}
\end{align}
This guarantees that the number of decomposed layers stays the same (two) after each decomposition step. 
This process is shown in Algorithm \ref{progressive-lrd}.
% \begin{figure}[t]
%   \includegraphics[width=0.5\linewidth]{figs/graphviz.png} \hfill
%   % \includegraphics[width=0.48\linewidth]{example-image-b}
%   \caption {Progressive decomposition scheme.}
%   \label{fig:pregressive}
% \end{figure}

% \begin{figure}[h]
%     \centering
%     \includegraphics[width=0.5\textwidth]{figs/LRD.pdf}
%     \caption{Progressive decomposition scheme}
%     \label{fig:pregressive}
% \end{figure}

\begin{algorithm}
\caption{Progressive LRD}
\label{progressive-lrd}
    % \REQUIRE $n \geq 0 \vee x \neq 0$
    % \ENSURE $y = x^n$
    \textbf{Input}: Pre-trained layer $\W$, Initial Rank $R_0$, and Rank reduction factor $0<\alpha<1$ \\
    \textbf{Output}: Decomposed layers $\W_0$ and $\W_1$ 
        
    \begin{algorithmic}[1]
    \STATE  $R \leftarrow R_0$ 
    \STATE $\W_0, \W_1 \leftarrow$ Decompose $\W$ using rank $R$ 
    \STATE $\W_0, \W_1 \leftarrow$ train $\W_0$ and $\W_1$
    % \STATE $T \leftarrow$ Decompose $\W$ using rank $R$ 
    % \STATE $R \leftarrow$ Rank for $2x$ compression
    % \STATE $R_{\min} \leftarrow$ Lower bound rank for $3x$ compression
    \WHILE{$R>1$ and desired compression ratio is not achieved}
        \STATE $R \leftarrow \alpha R$ 
        \STATE $\W_1^0, \W_1^1 \leftarrow$ Decompose $\W_1$ using rank $R$
        \STATE $\W_0 \leftarrow \W_0 \W_1^0$
        \STATE $\W_1 \leftarrow \W_1^1$
        \STATE $\W_0, \W_1 \leftarrow$ train $\W_0$ and $\W_1$
            % \IF{$r$ < $R$}
            %     \STATE $\Delta t(r) \leftarrow t(r)-t(r-1)$
            % \ENDIF
        % \STATE $r \leftarrow r-1$
    \ENDWHILE
    % \STATE \textbf{Derivative:} $\Delta t_r \leftarrow t_r-t_{r-1}$
    % \STATE \textbf{Optimal Rank:} $R_{opt} \leftarrow \argmax\limits_{r \in [R_{\min}, R]} \Delta t(r)$
    % \IF{$t(R_{opt})$ < $T$}
    %     \STATE Replace $L$ with $L_r$ 
    % \ELSE
    %     \STATE Use original layer $L$
    % \ENDIF
    \end{algorithmic}
\end{algorithm}

\section{Experiments} \label{experiment}
\subsection{Dataset}
SlimPajama-627B \cite{cerebras2023slimpajama} is a cleaned and deduplicated version of RedPajama \cite{together2023redpajama} dataset which is open-source. Table \ref{slim} shows the proportions of different data sources in the SlimPajama dataset. For our experiments, 1B of tokens are chosen from the first chunk of the dataset. As the dataset is shuffled before chunking, each chunk includes samples from different sources. This subset is further divided into 4 chunks of 250M tokens that are used in 4 steps of training after compression in PLRD experiments.
\begin{table}[h]
  \centering
  \begin{tabular}{cc}
    \hline
    \textbf{\small{Data Source}} & \textbf{\small{Proportion (\%)}} \\
    \hline
    \small{CommonCrawl} & \small{52.2\%} \\
    \small{C4} & \small{26.7\%} \\
    \small{GitHub} & \small{5.2\%} \\
    \small{Books} & \small{4.2\%} \\
    \small{ArXiv} & \small{4.6\%} \\
    \small{Wikipedia} & \small{3.8\%} \\
    \small{StackExchange} & \small{3.3\%} \\

    \hline
  \end{tabular}
  \caption{\label{slim}
    Proportions of data from different sources in SlimPajama-627B.  
  }
\end{table}

\subsection{PLRD Training}
LLaMa2 \cite{touvron2023llama} and Mistral-v0.1 \cite{jiang2023mistral} are two open-source model on which the efficacy of PLRD method is demonstrated. In each PLRD step, depending on if the compression includes attention matrices, MLP matrices, or both, intermediate ranks, $R_{attn}$ and $R_{MLP}$, are chosen. To scale each base model to 3B parameters, 4 steps of compression for each model are manually designed. In each step, the model is continually pretrained with 250M tokens to recover the performance.
These intermediate ranks were used to compress every matrix in attention or MLP module in every layer. The only exception is in Mistral-v0.1 where due to MQA technique in attention, the compression is only applied to $Q$ and $O$ matrices in attention modules and $K$ and $V$ matrices are untouched.
Table \ref{plrd} shows the intermediate ranks for each step of the compression for each model.
\begin{table}[h]
  \centering
  \begin{tabular}{ccc}
    \hline
    \textbf{\small{Model}} & \textbf{\small{$R_{attn}$}} & \textbf{\small{$R_{MLP}$}} \\
    \hline
    \small{PLRD-Mistral-v0.1-5.2B} & \small{NA} & \small{2048} \\
    \small{PLRD-Mistral-v0.1-4.9B} & \small{1536} & \small{2048} \\
    \small{PLRD-Mistral-v0.1-4.1B} & \small{1536} & \small{1536} \\
    \small{PLRD-Mistral-v0.1-3.1B} & \small{1536} & \small{1024} \\
    \hline
    \small{PLRD-LLaMa2-5.2B} & \small{NA} & \small{2048} \\
    \small{PLRD-LLaMa2-4.8B} & \small{1536} & \small{2048} \\
    \small{PLRD-LLaMa2-4.1B} & \small{1536} & \small{1536} \\
    \small{PLRD-LLaMa2-3.3B} & \small{768} & \small{1536} \\

    \hline
  \end{tabular}
  \caption{\label{plrd}
    Intermediate ranks in attention and MLP matrices in each step of compression.  
  }
\end{table}\subsection{PLRD Intermediate Results}
Intermediate results of the models after each step of the compression are reflected in tables \ref{mistral-plrd} and \ref{llama-plrd}. These results demonstrate that scaling models into different sizes with lightweight training is possible through PLRD method. The trade-off between size and model size shows that one can tailor open-source LLMs for a specific target computation budget and accuracy.
\begin{table*}
  \centering
  \begin{tabular}{cccccccccc}
    \hline
    \textbf{Model} \small{(\#Tokens to Train)} & \textbf{LogiQA} & \textbf{BoolQ} & \textbf{MMLU}&  \textbf{WinoGrande} & \textbf{Average}  \\
    \hline
    Mistral-v0.1-7B \small{(NA)} & 30.3\% & 83.6\% & 59.6\% & 74.0\% & 61.9\% \\  
    % LLaMA2-7B & 30.3\% & 77.8\% & 41.9\% & 69.0\% & 54.7\% \\
    \hline
    % OPT-2.7B & 25.8\% & 60.3\% & 25.7\% &60.9\% & 43.2\% \\
    % Open-LLaMA-3B-v2 & 28.0\% & 65.7\% & 25.5\% & 63.3\% & 45.6\% \\
    % Sheared-LLaMA-2.7B & 28.9\%^{\ast} & 66.0\% & 26.6\% & 64.2\%^{\ast} & 46.4\% \\
    % \hline
    PLRD-Mistral-v0.1-5.2B \small{(250M)} & 26.4\% & 72.6\% & 30.4\% & 66.8\% & 49.0\% \\
    PLRD-Mistral-v0.1-4.9B \small{(500M)}& 28.3\% & 69.6\% & 30.9\% & 63.3\% & 48.0 \\
    PLRD-Mistral-v0.1-4.1B \small{(750M)}& 28.9\% & 68.6\% & 27.3\% & 62.1\% & 46.7\%\\
    PLRD-Mistral-v0.1-3.1B \small{(1B)}& 26.9\% & 66.0\% & 26.5\% & 63.1\% & 45.6\%\\
    \hline
    
  \end{tabular}
  \caption{\label{mistral-plrd}
    Accuracy of each model in PLRD of Mistral-v0.1-7B base model. In each step of compression the model is continually pretrained with $250M$ tokens. The PLRD-Mistral-v0.1-3.1B model is trained with $1B$ tokens in total of 4 compression steps.
  }
\end{table*}
%--------------------
\begin{table*}
  \centering
  \begin{tabular}{cccccccccc}
    \hline
    \textbf{Model} \small{(\#Tokens to Train)} & \textbf{LogiQA} & \textbf{BoolQ} & \textbf{MMLU}&  \textbf{WinoGrande} & \textbf{Average}  \\
    \hline
    % Mistral-v0.1-7B & 30.3\% & 83.6\% & 59.6\% & 74.0\% & 61.9\% \\  
    LLaMA2-7B \small{(NA)}& 30.3\% & 77.8\% & 41.9\% & 69.0\% & 54.7\% \\
    \hline
    % OPT-2.7B & 25.8\% & 60.3\% & 25.7\% &60.9\% & 43.2\% \\
    % Open-LLaMA-3B-v2 & 28.0\% & 65.7\% & 25.5\% & 63.3\% & 45.6\% \\
    % Sheared-LLaMA-2.7B & 28.9\%^{\ast} & 66.0\% & 26.6\% & 64.2\%^{\ast} & 46.4\% \\
    % \hline
    PLRD-LLaMa2-5.2B \small{(250M)}& 26.4\% & 72.6\% & 30.4\% & 66.8\% & 49.0\%\\
    PLRD-LLaMa2-4.8B \small{(500M)}& 28.9\% & 69.1\% & 27.9\% & 64.7\% & 47.7\% \\
    PLRD-LLaMa2-4.1B \small{(750M)}& 28.9\% & 66.6\% & 26.5\% & 63.0\% & 46.3\%\\
    PLRD-LLaMa2-3.3B \small{(1B)}& 28.3\% & 64.1\% & 25.7\% & 63.4\% & 45.4\%\\
    \hline
    
  \end{tabular}
  \caption{\label{llama-plrd}
    Accuracy of each model in PLRD of LLaMa2-7B base model. In each step of compression the model is continually pretrained with $250M$ tokens. The PLRD-LLaMa2-3.3B model is trained with $1B$ tokens in total of 4 compression steps.
  }
\end{table*}
\subsection{Evaluation on Benchmarks}
For evaluation, lm-evaluation-harness \cite{gao2021framework} package is used. Performance of the models on 4 diverse downstream tasks, LogiQA, BoolQ, MMLU, and WinoGrande, are evaluated in zero-shot manner. According to table \ref{benchmark-3b}, evaluations show that PLRD-Mistral-v0.1-3.1B and PLRD-LLaMa2-3.3B achieve on par performance with other open-source models with comparable size while trained on only 1B tokens. The results of PLRD are similar for both Mistral-v0.1-3.1B and LLaMa2-3.3B which yields the conclusion that PLRD is a general method applicable to a variety of models. 
\subsection{Inference Speed}
As PLRD changes the depth of the model by replacing a matrix with two low-rank matrices, one reasonable investigation is on the inference speed of the PLRD models compared to the conventional architectures of LLMs. To analyze the inference speed of the models, generation speed of PLRD-Mistral-v0.1-3.1B on CPU is compared with a similar size model with conventional architecture in Table \ref{infer-speed}. The analysis shows that PLRD model with similar size is less than 3\% slower than the conventional model.
\begin{table}[h]
  \centering
  \begin{tabular}{cc}
    \hline
    \textbf{\small{Model}} & \textbf{\small{Inference Speed (T/s)}} \\
    \hline
    
    \small{Open-LLaMA-3B-v2} & \small{5.40} \\
    \small{PLRD-Mistral-v0.1-3.1B} & \small{5.24 \textbf{(-2.96\%)}}\\

    \hline
  \end{tabular}
  \caption{\label{infer-speed}
    Inference Speed Analysis on CPU. The inference speed of PLRD-Mistral-v0.1-3.1B is only $2.96\%$ slower compared to Open-LLaMA-3B-v2. 
  }
\end{table}

\subsection{Experimental Setup}
For our experiments, we used the Mistral-v0.1-7B and LLaMa2-7B publicly available on HuggingFace as base models. In each step of compression, the model is continually pretrained with 250M tokens. 8 V100 gpus were used all experiments in the paper. As the purpose of this paper is explore a new compression technique and not get optimal results from this technique, no hyperparameter search is done. One set of parameters were used for all the experiments which are listed in Table \ref{hyperparameter}.
\begin{table}[h]
  \centering
  \begin{tabular}{cc}
    \hline
    \textbf{Parameter} & \textbf{Value}\\
    \hline
    Optimizer &AdamW\\
    Warmup Ratio &0.03\\
    LR Scheduler & Linear\\
    Epoch & 1\\
    Learning Rate&2e-5\\
    Weight Decay & 0\\
    Max. Seq. Len. & 512\\

    \hline
  \end{tabular}
  \caption{\label{hyperparameter}
    Hyperparameters used throughout all experiments in this paper. As the purpose of this paper is not to identify best hyperparameters, no hyperparameter tuning is done.
  }
\end{table}

\section{Conclusion}
The method of Progressive Low-rank Decomposition (PLRD), when used to scale open-source LLMs, demonstrates better or comparable results with models of similar size that are pre-trained from scratch.
Compared to our other compression methods such as Sheared-LLaMa \cite{xia2023sheared}, PLRD achieves similar results without the need to use tailored training recipe or a large number of training tokens.
The results demonstrate that PLRD is effective in scaling down the models. As there is no specific modification in the training and the dataset selection is random, the experimental results support the claim that this method is a general method for compressing LLMs. Also, other methods that improve the accuracy of the models in training such as knowledge distillation, dynamic batch loading \cite{xia2023sheared}, etc. are orthogonal to our method and can further improve the results.
\section{Limitations and Future Work}
Our approach depends on a pre-trained open-source model. Due to limited time and computation resources, only two models with 7B parameters were chosen for experiments. However, our approach leverages the redundancy in the model parameters and our intuition is that PLRD will show stronger results if applied to larger models, which would be the next step of this project.
\section{Ethics Statement}
It should be noted that the models trained in this paper are trained on open-source datasets without further safety measures to align the model generations with ethical rules. The models are evaluated on benchmarks that test the model's capacity to reason. Therefore, models' generations might include false information, bias and discrimination, etc. It is not advised to use these models for purposes other than research.
\bibliography{acl_latex}

\appendix

\section{Appendix}
\label{sec:appendix}

\subsection{Comparative Analysis of Qualitative Response Generation by Language Models}
In Table \ref{table-example}, we selected a random set of prompts from Alpaca dataset \cite{alpaca} and used Open-LLaMA-3B-v2 and PLRD-Mistral-v0.1-3.1B to generate responses. Comparing model generated responses show that  after the proposed model compressing approach, PLRD-Mistral-v0.1-3.1B after continual pretraining, is able to generate Produce responses that are equally coherent and informative, and even better than Open-LLaMA-3B-v2 in some cases.

\begin{table*}[h]
% \label{"}
\centering
\begin{tabular}{|>{\centering\arraybackslash}m{0.5\textwidth}|>{\centering\arraybackslash}m{0.5\textwidth}|}
\hline
\textbf{Open-LLaMA-3B-v2} & \textbf{PLRD-Mistral-v0.1-3.1B} \\ \hline
\multicolumn{2}{|>{\raggedright\arraybackslash}m{1\textwidth}|}{\textbf{Query:} Propose an idea for a self-driving car.} \\ \hline
\multicolumn{1}{|>{\raggedright\arraybackslash}m{0.5\textwidth}|}{I think that the self-driving car should be able to detect and avoid obstacles.} & \multicolumn{1}{>{\raggedright\arraybackslash}m{0.5\textwidth}|}{A self-driving car is a vehicle that can drive itself without the driver having to be present.} \\ \hline
\multicolumn{2}{|>{\raggedright\arraybackslash}m{1\textwidth}|}{\textbf{Query:} Describe how to prepare Chewy Chocolate Chip Cookies in your own words.} \\ \hline
\multicolumn{1}{|>{\raggedright\arraybackslash}m{0.5\textwidth}|}{I will need to use a mixing bowl, a spoon, and a baking sheet. I will need to mix the flour, baking soda, and salt together. I will need to add the butter, brown sugar, and white sugar to the bowl
. I will need to mix the ingredients together until they are well combined. I will need to add the eggs and vanilla extract to the bowl.
} & \multicolumn{1}{>{\raggedright\arraybackslash}m{0.5\textwidth}|}{1. Prepare the oven for 20 minutes. \newline
        2. Prepare the baking sheet for the cookies. \newline
        3. In a bowl, combine the ingredients for the cookies.\newline
        4. Pour the mixture into the prepared baking sheet.\newline
        5. Bake the cookies for 20 minutes.\newline
        6. Remove the cookies from the oven and let cool.\newline
        7. Store the cookies in an airtight container.\newline} \\ \hline
\multicolumn{2}{|>{\raggedright\arraybackslash}m{1\textwidth}|}{\textbf{Query:} Explain the concept of cogging torque.} \\ \hline
\multicolumn{1}{|>{\raggedright\arraybackslash}m{0.5\textwidth}|}{Cogging torque is a torque that is induced in a motor due to the back EMF of the motor.
        The back EMF is the voltage that is induced in the motor when the motor is rotating.
} & \multicolumn{1}{>{\raggedright\arraybackslash}m{0.5\textwidth}|}{Cogging torque is the torque that is produced when the wheel is turning.} \\ \hline
\multicolumn{2}{|>{\raggedright\arraybackslash}m{1\textwidth}|}{\textbf{Query:} Explain the concept of artificial intelligence in simple terms.} \\ \hline
\multicolumn{1}{|>{\raggedright\arraybackslash}m{0.5\textwidth}|}{Artificial intelligence is a branch of computer science that deals with the creation of intelligent machines.} & \multicolumn{1}{>{\raggedright\arraybackslash}m{0.5\textwidth}|}{Artificial intelligence is a branch of computer science that deals with the creation of computer programs that can perform tasks that humans cannot. These programs are designed to be able to perform tasks that humans cannot do.} \\ \hline
\end{tabular}
\caption{\label{table-example} Generated responses by  Open-LLaMA-3B-v2 and PLRD-Mistral-v0.1-3.1B models for a set of sample prompts from alpaca dataset. Comparative analysis reveals equivalence in the quality of responses generated by the two language models.}
\end{table*}

\end{document}